\documentclass{article} 
\usepackage{iclr2015,times}
\usepackage{hyperref}
\usepackage{url}

\usepackage{amsmath,amsbsy,amsfonts,amssymb,amsthm,dsfont,units,psfrag}
\usepackage{graphicx}
\usepackage{algorithm,algorithmic,mathtools}
\usepackage{color,cases}

\usepackage{tikz}
\usetikzlibrary{calc, shapes, backgrounds, arrows, fit, calc}
\usepackage{subfigure}
\usepackage{hyperref}
\usepackage{appendix}

\def\Pc{{\cal S}}

\newcommand\tl{\tilde}

\newcommand\blfootnote[1]{%
  \begingroup
  \renewcommand\thefootnote{}\footnote{#1}%
  \addtocounter{footnote}{-1}%
  \endgroup
}

\def\tl{\tilde}

 \def\0{{\bf 0}}

\def\qed{\hfill\hbox{${\vcenter{\vbox{
    \hrule height 0.4pt\hbox{\vrule width 0.4pt height 6pt
    \kern5pt\vrule width 0.4pt}\hrule height 0.4pt}}}$}}

\definecolor{myred}{rgb}{0.3,0.0,0.7}
\definecolor{dkg}{rgb}{0.1,0.7,0.2}
\definecolor{dkb}{rgb}{0.0,0.2,0.8}

\def\Ebb{{\mathbb E}}

\newcommand{\bprf}{\begin{myproof}}
\newcommand{\eprf}{\end{myproof}}
\newcommand{\bp}{\begin{psfrags}}
\newcommand{\ep}{\end{psfrags}}
\newcommand{\bl}{\begin{lemma}}
\newcommand{\el}{\end{lemma}}
\newcommand{\bt}{\begin{theorem}}
\newcommand{\et}{\end{theorem}}
\newcommand{\bc}{\begin{center}}
\newcommand{\ec}{\end{center}}
\newcommand{\bi}{\begin{itemize}}
\newcommand{\ei}{\end{itemize}}
\newcommand{\ben}{\begin{enumerate}}
\newcommand{\een}{\end{enumerate}}
\newcommand{\bd}{\begin{definition}}
\newcommand{\ed}{\end{definition}}
\def\beq{\begin{equation}}
\def\eeq{\end{equation}\noindent}
\def\beqn{\begin{eqnarray}}
\def\eeqn{\end{eqnarray} \noindent}
\def\beqnn{  \begin{eqnarray*}}
\def\eeqnn{\end{eqnarray*}  \noindent}
\def\bcase{  \begin{numcases}}
\def\ecase{\end{numcases}   \noindent}
\def\bsbcase{  \begin{subnumcases}}
\def\esbcase{\end{subnumcases}   \noindent}

\newtheorem{theorem}{Theorem}

\newtheorem{lemma}[theorem]{Lemma}

\newtheorem{definition}{Definition}

\newenvironment{myproof}{\noindent{\bf Proof:} \hspace*{1em}}{
    \hspace*{\fill} $\Box$ }
\newenvironment{proof_of}[1]{\noindent {\bf Proof of #1: }}{\hspace*{\fill} $\Box$ }

\newcommand{\matplottc}[1]{               
        \unitlength .45truein
        \begin{center}
        \includegraphics{#1.ps}
        \end{picture}
        \end{center}
}

\def\psfancypar#1#2{\begingroup\def\par{\endgraf\endgroup\lineskiplimit=0pt}
               \setbox2=\hbox{\large\sc #2}
               \newdimen\tmpht \tmpht \ht2 \advance\tmpht by \baselineskip
               \font\hhuge=Times-Bold at \tmpht
               \setbox1=\hbox{{\hhuge #1}}
               \count7=\tmpht \count8=\ht1
               \divide\count8 by 1000 \divide\count7 by \count8
               \tmpht=.001\tmpht\multiply\tmpht by \count7
               \font\hhuge=Times-Bold at \tmpht
               \setbox1=\hbox{{\hhuge #1}}
               \noindent
                \hangindent1.05\wd1
               \hangafter=-2 {\hskip-\hangindent
               \lower1\ht1\hbox{\raise1.0\ht2\copy1}%
                \kern-0\wd1}\copy2\lineskiplimit=-1000pt}

\def\Kout{\setbox1=\hbox{\Huge\bf K}\hbox to
1.05\wd1{\hspace{.05\wd1}
}}
\def\Sout{\setbox1=\hbox{\Huge\bf S}\hbox to 1.05\wd1{\hspace{.05\wd1}
}}

\newcommand{\torestate}[3]{%
\expandafter \def \csname BBRESTATE #2 \endcsname{#3}
\theoremstyle{plain}
\newtheorem{BBRESTATETHMNUM#2}[theorem]{#1}
\begin{BBRESTATETHMNUM#2}\label{#2}\csname BBRESTATE #2 \endcsname   \end{BBRESTATETHMNUM#2}
\newtheorem*{BBRESTATETHMNONNUM#2}{{#1}~\ref{#2}}
}

\newcommand{\restate}[1]{\begin{BBRESTATETHMNONNUM#1}[Restated] \csname BBRESTATE #1 \endcsname
\end{BBRESTATETHMNONNUM#1}}

\title{Score Function Features for Discriminative Learning}

\author{
Majid Janzamin \\
Department of Electrical Engineering and Computer Science \\
University of California \\
Irvine, CA 92697, USA \\
\texttt{mjanzami@uci.edu}
\And
Hanie Sedghi\\
Department of Electrical Engineering\\
University of Southern California \\
Los Angeles, CA 90089, USA \\
\texttt{hsedghi@usc.edu}
\And
Anima Anandkumar \\
Department of Electrical Engineering  and Computer Science \\
University of California \\
Irvine, CA 92697, USA \\
\texttt{a.anandkumar@uci.edu}
}

\iclrfinalcopy 

\begin{document}

\maketitle

\begin{abstract}
Feature learning forms the cornerstone for tackling challenging learning problems in domains such as speech, computer vision and natural language processing. In this paper, we consider a novel class of matrix and tensor-valued features, which can be pre-trained using unlabeled samples. We present efficient algorithms for extracting discriminative information, given these pre-trained features and labeled samples for any related task. Our class of   features are based on higher-order score functions, which capture local variations in the probability density function of the input. We establish a theoretical framework to characterize the nature of discriminative information that can be extracted from score-function features, when used in conjunction with labeled samples.  We employ efficient spectral decomposition algorithms (on matrices and tensors) for extracting discriminative components. The advantage of employing tensor-valued features is that we can extract richer discriminative information in the form of an overcomplete representations. Thus, we present a novel framework for employing   generative models of the input for discriminative learning.
\end{abstract}

\paragraph{Keywords:}Feature learning, semi-supervised learning, self-taught learning, pre-training, score function,  spectral decomposition methods, tensor methods. \blfootnote{A longer version of this work is available on arXiv: \url{http://arxiv.org/abs/1412.2863}.}

\section{Introduction}

Having good features or representations of the input data  is critical to achieving good performance in challenging machine learning tasks in domains such as speech, computer vision and natural language processing~\citep{bengio2013representation}. Traditionally, feature  engineering   relied on carefully hand-crafted features,  tailored towards a specific task: a laborious and a time-consuming process.
Instead, the recent trend has been to automatically learn good features through various frameworks such as deep learning~\citep{bengio2013representation}, sparse coding~\citep{raina2007self}, independent component analysis (ICA)~\citep{le2011ica}, Fisher kernels~\citep{jaakkola1999exploiting}, and so on. These approaches are unsupervised and can thus exploit the vast amounts of unlabeled samples, typically present in these domains.

A good feature representation  incorporates important prior knowledge about the input, typically through a probabilistic model. In almost every conceivable scenario, the probabilistic model needs to incorporate latent variables to fit the input data. These latent factors can be important explanatory variables for classification  tasks associated with the input. Thus, incorporating generative models of the input can hugely boost the performance of discriminative tasks.

Many approaches to feature learning  focus   on unsupervised  learning, as described above.
The hypothesis behind employing unsupervised  learning is  that the input distribution is related to the associative model between the input and the label of a given task, which is reasonable to expect in most scenarios.  When the distribution of the unlabeled samples, employed for feature learning, is the same as the labeled ones, we have the framework of   {\em semi-supervised} learning. A more general framework, is the so-called {\em self-taught} learning, where the distribution of unlabeled samples is different, but related to the labeled ones~\citep{raina2007self}. Variants of these frameworks include transfer learning, domain adaptation and multi-task learning~\citep{bengio2012deep}, and involve labeled datasets for related tasks.  These frameworks have been of extensive interest to the machine learning community, mainly  due to the scarcity of labeled samples for many challenging tasks. For instance,  in   computer vision, we have a huge corpus of unlabeled images, but a more limited set of labeled ones. In natural language processing, it is extremely laborious to annotate the text with syntactic and semantic parses, but we have access to unlimited amounts of unlabeled text.

%


It has been postulated that humans mostly learn in an unsupervised manner~\citep{raina2007self}, gathering ``common-sense'' or ``general-purpose'' knowledge, without worrying about any specific goals. Indeed, when faced with a specific task, humans  can quickly and easily extract  relevant   information  from the accrued  general-purpose knowledge.  Can we design machines with similar capabilities? Can we design algorithms which  succinctly summarize  information in  unlabeled samples as  general-purpose features? When given a specific task, can we efficiently extract relevant  information from general-purpose features? Can we provide  theoretical guarantees for such algorithms? These are indeed challenging questions, and we   provide some concrete answers  in this paper.


\section{Summary of Results}

In this paper, we consider a class of matrix and tensor-valued ``general-purpose'' features,  pre-trained  using unlabeled samples. We assume that the labels are not present at the time of feature learning. When presented with labeled samples, we leverage these pre-trained features to  extract discriminative information using efficient spectral decomposition algorithms. As a main contribution, we provide theoretical guarantees on the nature of discriminative information that can be extracted with our approach.


We consider the class of features based on higher-order score functions of the input, which involve higher-order derivatives of the   probability density function (pdf). These functions    capture  ``local manifold structure'' of the pdf. While the first-order score function is a vector (assuming a vector input), the higher-order functions are matrices and tensors, and thus capture richer information about the input distribution.
Having access to these matrix and tensor-valued features allows to extract better discriminative information, and we characterize its precise nature in this work.

Given  score-function features and  labeled samples,   we extract discriminative information  based on the   method of moments. We construct cross-moments involving the labels and the input score features. Our main theoretical result is  that these moments are equal to the expected derivatives of the label, as a function of the input or some model parameters. In other words, these moments capture  variations of the label function, and are therefore informative for discriminative tasks.

We   employ  spectral decomposition algorithms  to find succinct representations of the moment matrices/tensors. These algorithms are fast  and embarrassingly parallel. See ~\citep{AnandkumarEtal:tensor12,JanzaminEtal:Altmin14,JanzaminEtal:Altmin14-2} for details, where we have developed and analyzed efficient tensor decomposition algorithms (along with our collaborators). The advantage of the tensor methods is that they do not suffer from spurious local optima, compared to typical non-convex problems such as expectation maximization or backpropagation in neural networks. Moreover,
we can construct overcomplete representations for tensors, where the number of  components in the representation can exceed the data dimensionality.
It has been  argued that having overcomplete   representations  is crucial to getting good classification performance~\citep{coates2011analysis}.
Thus, we can leverage the latest advances in spectral methods for efficient extraction of  discriminative information from moment  tensors.

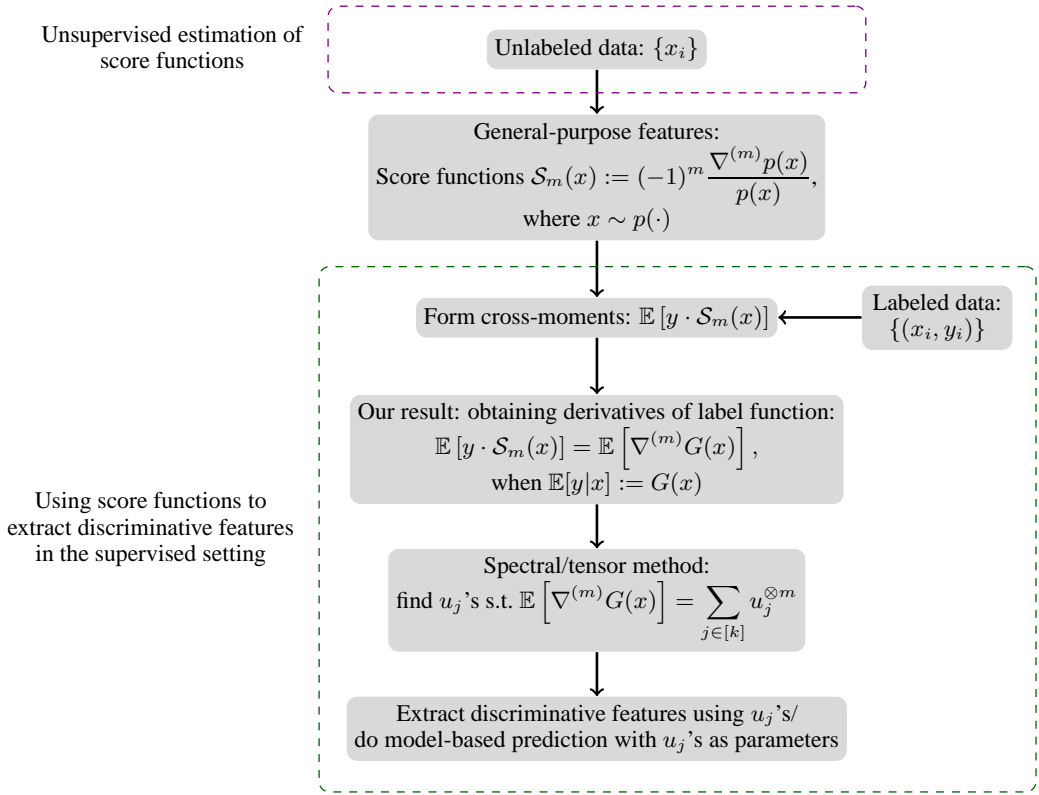
\begin{figure}[t]
\bc
\begin{tikzpicture}
[
scale=1,
   nodestyle/.style={fill = gray!30, shape = rectangle, rounded corners, minimum width = 2cm},
]
\small
\matrix [column sep=2mm,row sep=6mm] {
\node[nodestyle](a){Unlabeled data: $\{x_i\}$}; & \\
\node[nodestyle, align=center](b){General-purpose features: \\ Score functions $\Pc_m(x) := (-1)^m \dfrac{\nabla^{(m)} p(x)}{p(x)}$, \\ where $x \sim p(\cdot)$}; & \\
\node[nodestyle, align=center](c){Form cross-moments: $\displaystyle \Ebb \left[y \cdot \Pc_m(x) \right]$}; & \node[nodestyle, align=center](c2){Labeled data: \\ $\{(x_i,y_i)\}$}; \\
\node[nodestyle, align=center](d){Our result: obtaining derivatives of  label function: \\ $\displaystyle \Ebb \left[y \cdot \Pc_m(x) \right] = \Ebb \left[ \nabla^{(m)} G(x) \right],$ \\ when $\Ebb[y|x]:=G(x)$ }; & \\
\node[nodestyle, align=center](e){Spectral/tensor method: \\ find $u_j$'s s.t.\ $\displaystyle\Ebb \left[ \nabla^{(m)} G(x) \right]= \sum_{j \in [k]} u_j^{\otimes m}$}; & \\
\node[nodestyle, align=center](f){Extract discriminative features using $u_j$'s/ \\ do model-based prediction with $u_j$'s as parameters}; & \\
};
\draw [->, line width = 1pt] (a) to (b);
\draw [->, line width = 1pt] (b) to (c);
\draw [->, line width = 1pt] (c) to (d);
\draw [->, line width = 1pt] (d) to (e);
\draw [->, line width = 1pt] (e) to (f);
\draw [->, line width = 1pt] (c2) to (c);

\node [draw, dashed, rounded corners, violet, line width=0.5pt,
	fit = {(a) ($(a.east)+(20mm,0)$) ($(a.west)-(20mm,0)$) ($(a.north)+(0,2mm)$) ($(a.south)-(0,2mm)$)},
	label=left:{\begin{tabular}{c} Unsupervised estimation of\\ score functions \end{tabular}}
	] {};

\node [draw, dashed, rounded corners, green!40!black, line width=0.5pt,
	fit = {(c) (c2) (d) (e) (f) ($(c2.east)+(2mm,0)$) ($(d.west)-(3mm,0)$) ($(c.north)+(0,3mm)$) ($(f.south)-(0,3mm)$)},
	label=left:{\begin{tabular}{c} Using score functions to \\ extract
discriminative features \\ in the supervised setting \end{tabular}}
	] {};

\end{tikzpicture}
\ec
\caption{\small Overview of the proposed framework of using the general-purpose features to generate discriminative features through spectral methods.}\label{fig:overview}
\end{figure}

In our framework, the  label can be a scalar,  a vector, a matrix or even a tensor, and it can either be continuous or discrete. We can therefore handle a variety of regression and classification settings such as multi-task, multi-class,   and structured prediction problems.
Thus, we present a unified and an efficient end-to-end framework for extracting discriminative information from pre-trained features. An overview of the entire framework is presented in Figure~\ref{fig:overview} which is fully explained later in Section~\ref{sec:overview}.

We now provide some important observations below.

\paragraph{Are the expected label function derivatives informative? }
Our analysis characterizes the discriminative information we can extract from score function features. As described above, we prove that  the cross-moments between the label and the score function features are equal to the expected derivative of the label as a function of the input or model parameters. But  when are these expected label derivatives informative? Indeed, in trivial cases, where the derivatives of the label function vanish over the  support of the input distribution, these moments carry no information. However, such cases are pathological, since then, either there is  no variation   in  the label function or the input distribution is nearly degenerate. Another possibility is that  a certain derivative vanishes, when averaged over the input distribution, even though it is not zero everywhere. If this is the case, then the next derivative cannot be averaged out to zero, and will thus carry information about the variations of the label function. Thus, in practical scenarios, the cross-moments contain useful discriminative information. In fact,  for many discriminative models which are challenging to learn, such as multi-layer neural networks and mixtures of classifiers, we establish that these moments have an intimate relationship with the parameters of the discriminative model in subsequent works~\citep{Sedghi:SparseNet,Sedghi:mixture}. Spectral decomposition of the moments provably recovers the model parameters. These are the first results for guaranteed  learning of many challenging discriminative latent variable models.

\paragraph{Contrasting with previous approaches: }We now contrast our approach to previous approaches for incorporating generative models in discriminative tasks. Typically, these approaches directly feed the pre-trained features   to a classifier. For example, in the Fisher kernel framework, the Fisher score features are fed to a kernel classifier~\citep{jaakkola1999exploiting}. The reasoning behind this is that the  features obtained from unsupervised learning have information  about all the classes, and the task of finding class-specific differences in the learnt representation is left to the  classifier. However, in practice, this may not be the case, and a common complaint is that these generative features are not discriminative for the task at hand. Previous solutions have prescribed joint training discriminative features using labeled samples, in conjunction with unlabeled samples~\citep{mairal2009supervised,maaten2011learning,wang2013robust}. However, the resulting optimization problems are complex and expensive to run, may not converge to good solutions, and have to be re-trained for each new task. We present an alternative approach to extract discriminative features using efficient spectral decomposition algorithms on   moment  matrices and tensors. These methods are light weight and fast,   and we theoretically quantify the nature of discriminative features they can extract. These discriminative features can then be fed into the classification pipeline.  Thus, the advantage of our approach is that we can quickly generate discriminative features for new classification tasks without going through the laborious process of re-training for   new features.

We now contrast our approach with previous moment-based approaches for discriminative learning, which   consider moments between the label and  raw input, e.g.~\citep{karampatziakis2014discriminative}. Such methods have no theoretical guarantees. In contrast, we construct cross-moments between the label and the score function features.   We show that using score function features   is crucial to mining discriminative information with provable guarantees.

\paragraph{Extension to self-taught learning: }
We have so far described our framework under the semi-supervised setting, where the unlabeled and labeled samples have the same input distribution. We can also handle the framework of self-taught learning, where the two  distributions are related but may not be the same.  We prescribe some simple pre-processing to transfer the parameters and to re-estimate the score function  features  for the input of the labeled data set. Such parameter transfer frameworks have been considered before, e.g. \citep{raina2007self}, except here we present a general latent-variable framework and  focus on transferring parameters for computing score functions, since we require them for subsequent operations. Our framework can also  be applied to   scenarios  where we  have different input sources with different distributions, but the classification task is the same, and thus, the associative model between the label and the input is fixed. Consider for instance,   crowdsourcing applications, where the same task is presented to different groups of individuals. In our approach, we can then construct different score function features for different input sources and the different cross-moments provide  information about the variations in the label function, averaged over different input distributions. We can thus leverage the diversity of different input sources for improved performance on  common tasks. Thus, our approach is applicable in many challenging practical scenarios. 

\section{Overview of the Framework} \label{sec:overview}

In   this section, we elaborate on the end-to-end framework presented in Figure~\ref{fig:overview}.

\paragraph{Background:} The problem of supervised learning consists of learning a predictor, given labeled training samples $\{(x_i, y_i)\}$ with input   $x_i$ and corresponding  label $y_i$. Classical frameworks such as SVMs are purely discriminative since they make no distributional assumptions. However, when labeled data is limited and classification tasks are challenging,  incorporating distributional information can improve performance.  In an associative  model-based framework, we   posit a conditional distribution for the label given the input $p(y|x)$. However, learning this model is challenging, since maximum-likelihood estimation of $p(y|x)$ is non-convex and NP-hard to solve in general, especially if it involves hidden variables (e.g., associative mixtures, multi-layer neural networks). In addition, incorporating a generative model for input $x$ often leads to improved   discriminative performance.

\paragraph{Label-function derivatives are discriminative:}
Our main focus in this work is to extract useful information about $p(y|x)$ without attempting to learn it in its entirety. In particular, we extract  information about the local variations of conditional distribution $p(y|x)$, as the  input $x$  (or some model parameter) is changed. For the classification setting, it suffices to consider\footnote{In the classification setting, powers of $y$, e.g., $y^2$ contain no additional information, and hence, all the information of the associative model is in $\Ebb[y|x]:=G(x)$. However, in the regression setting, we can compute additional functions, e.g., $\Ebb[\nabla^{(m)} H(x)]$, where $\Ebb[y^2 | x] := H(x)$. Our approach can also compute these derivatives.}   $\Ebb[y|x]:=G(x)$. In this paper, we present mechanisms to estimate its expected higher order derivatives\footnote{Note that since we are computing the expected derivatives, we also assume a distribution for  the input $x$.}
\beq \label{eqn:derivative-intro} \Ebb[\nabla_x^{(m)} G(x)], \ m \geq 1,\eeq
where $\nabla^{(m)}_x$ denotes the $m$-th order derivative operator w.r.t.\ variable $x$.
By having access to expected derivatives of the label function   $G(x)$ in \eqref{eqn:derivative-intro}, we   gain an understanding of how the label $y$ varies  as we   change the input $x$ locally, which is valuable discriminative information.


\paragraph{Score functions yield  label-function derivatives:}
One of the main contributions of this paper is to obtain these expected derivatives  in \eqref{eqn:derivative-intro} using features denoted by  $\Pc_m (x)$, for $m \geq 1$ (learnt from unlabeled samples) and the labeled data. In particular, we form the cross-moment between the label $y$ and the features $\Pc_m(x)$, and show that they yield the derivatives as\footnote{We drop subscript $x$ in the derivative operator $\nabla^{(m)}_x$ saying $\nabla^{(m)}$ when there is no ambiguity.}
\beq \Ebb[ y \cdot \Pc_m(x)] =   \Ebb[\nabla^{(m)} G(x)], \quad \mbox{when }\Ebb[y|x]:= G(x).\label{eqn:yield}\eeq

We establish a simple form  for features  $\Pc_m(x)$,  based on the derivatives of the probability density function $p(\cdot)$ of the input $x$ as
\beq \label{eqn:highorderintro}\Pc_m(x) = (-1)^m\frac{\nabla^{(m)} p(x)}{p(x)}, \quad \mbox{when } x \sim p(\cdot). \eeq
In fact, we show that the feature $\Pc_m(x)$ defined above is a function of higher order score functions $\nabla_x^{(n)} \log p(x)$ with $n \leq m$, and we derive an  explicit relationship between them. This is basically why we also call these features as (higher order) score functions. Note that the features $\Pc_m(x)$ can be learnt using unlabeled samples, and we term them  as general-purpose features since they can be applied to any labeled dataset, once they are estimated. Note the features $\Pc_m(x)$ can be   vectors, matrices or  tensors, depending on $m$, for multi-variate $x$. The choice of order $m$ depends on the particular setup: a higher $m$ yields more information (in the form of higher order derivatives) but requires more samples to compute the empirical moments accurately.

We then extend the framework to parametric setting, where we obtain derivatives $\Ebb[\nabla^{(m)}_\theta G(x;\theta)]$ with respect to some model parameter $\theta$ when $\Ebb[y|x;\theta]:=G(x;\theta)$. These are obtained using general-purpose features denoted by $\Pc_m(x;\theta)$ which  is a function of higher order Fisher score functions $\nabla_\theta^{(n)} \log p(x;\theta)$ with $n \leq m$. Note that by  using the parametric framework   we can now incorporate discrete input $x$, while this is not possible with the previous framework.

\paragraph{Spectral decomposition of derivative matrices/tensors:}
Having obtained the derivatives $\Ebb[\nabla^{(m)} G(x)]$  (which are  matrices or tensors), we then find efficient representations  using spectral/tensor decomposition  methods. In particular, we find vectors $u_j$ such that \beq\label{eqn:decomp} \Ebb[\nabla^{(m)} G(x)] =\sum_{j \in [k]} \overbrace{u_j\otimes u_j\otimes \cdots \otimes u_j}^{m \mbox{ times}},\eeq where $\otimes$ refers to the tensor product notation. Note that since the higher order derivative is a symmetric matrix/tensor, the decomposition is also symmetric. Thus, we decompose the matrix/tensor at hand into sum of rank-$1$ components, and in the matrix case, this reduces to computing the SVD.  In the case of a tensor, the above decomposition is termed as CP decomposition~\citep{Kruskal:77}. In a series of works~\citep{AnandkumarEtal:tensor12,JanzaminEtal:Altmin14,JanzaminEtal:Altmin14-2}, we have presented efficient algorithms for obtaining \eqref{eqn:decomp}, and analyzed their performance in detail. 


The matrix/tensor in hand is decomposed into a sum of $k$ rank-1 components. Unlike matrices,  for tensors, the  rank parameter $k$ can be larger than the dimension. Therefore, the decomposition problems falls  in to two different regimes. One is the undercomplete regime: where $k$ is less than the dimension, and the overcomplete one, where it is not. The undercomplete regime  leads to  dimensionality reduction, while the overcomplete regime results in richer representation. 

Once we obtain components $u_j$, we then have several options to perform further processing. We can extract discriminative features such as $\sigma(u_j^\top x)$, using some non-linear function $\sigma(\cdot)$, as performed in some of the earlier works, e.g.,~\citep{karampatziakis2014discriminative}.
Alternatively, we can perform model-based prediction and incorporate $u_j$'s as parameters of a discriminative model.  In a subsequent paper, 
we show that $u_j$'s correspond to significant   parameters of many challenging discriminative models such as multi-layer feedforward neural networks and mixture of classifiers, under the {\em realizable} setting.

\paragraph{Extension to self-taught learning: }
The results presented so far assume the semi-supervised setting, where the unlabeled samples $\{\tilde{x}_i\}$ used to estimate the score functions are drawn from the same distributions as the  input $\{x_i\}$ of the labeled samples $\{(x_i, y_i)\}$. We present simple mechanisms to extend to the self-taught setting, where the distributions of $\{\tilde{x}_i\}$ and $\{x_i\}$ are related, but not the same. We assume latent-variable models for $\tilde{x}$ and $x$, e.g., sparse coding, independent component analysis (ICA), mixture models, restricted Boltzmann machine (RBM), and so on. We assume that the conditional distributions $p(\tilde{x}| \tilde{h})$ and $p(x|h)$, given the corresponding latent variables $\tilde{h}$ and $h$ are the same. This is reasonable since the unlabeled samples $\{\tilde{x}_i\}$ are usually ``rich'' enough to cover all the elements. For example,  in the sparse coding setting, we assume that all the dictionary elements can be learnt through $\{\tilde{x}_i\}$, which is assumed in a number of previous works, e.g~\citep{raina2007self,zhang2008flexible}. Under this assumption, estimating the score function for new samples $\{x_i\}$ is relatively straightforward, since we can transfer the estimated conditional distribution $p(\tl{x}|\tl{h})$ (using unlabeled samples $\{\tl{x}_i\}$) as the estimation of $p(x|h)$, and we can re-estimate the marginal distribution $p(h)$ easily. Thus, the use of  score functions allows for easy transfer of information under the self-taught framework. The rest of the steps can proceed as before.

\section{Conclusion}
We provided a general framework for proposing discriminative features by introducing higher order score functions. The framework can be applied in semi-supervised and self-taught learning settings. We first use the unlabeled data to estimate the score function. We then show that the score function yields the expected label-function derivative by forming the cross-moment between the label and the score function. Then we use spectral methods to extract discriminative features by decomposing the higher order derivatives of label-function into rank-1 components. We apply this framework for learning several challenging models such as multi-layer neural networks and mixtures of classifiers in subsequent works~\citep{Sedghi:SparseNet,Sedghi:mixture}.

\subsubsection*{Acknowledgments}

M. Janzamin thanks Rina Panigrahy for useful discussions.
M. Janzamin is supported by NSF Award CCF-1219234. H. Sedghi is supported by ONR Award N00014-14-1-0665. A. Anandkumar is supported in part by Microsoft Faculty Fellowship, NSF Career award CCF-$1254106$, NSF Award CCF-$1219234$, ARO YIP Award W$911$NF-$13$-$1$-$0084$ and ONR Award N00014-14-1-0665.


\end{document}